\PassOptionsToPackage{table}{xcolor}
\documentclass[10pt,twocolumn,letterpaper]{article}
\usepackage{iftex}
\ifXeTeX\usepackage[T1]{fontenc}\fi
\usepackage[pagenumbers]{cvpr} 
\usepackage{cuted}
\usepackage{currfile}
\usepackage{caption}
\usepackage{graphicx}
\usepackage{booktabs}
\usepackage{adjustbox}
\usepackage{multirow}
\usepackage{amsmath,amssymb}
\usepackage{colortbl}
\usepackage{xspace}
\usepackage{pifont}
\usepackage{microtype}
\usepackage{placeins}
\definecolor{oursred}{HTML}{8B1A1A}
\definecolor{oursbg}{HTML}{FBF3F3}
\newcommand{\ours}{\textsc{Margin}\xspace}
\newcommand{\aspire}{ASPIRE\xspace}

\newcommand{\Sdev}{\mathcal{S}_{\mathrm{dev}}}
\newcommand{\Tmax}{T_{\max}}

\newcommand{\nCostCampTaskMax}{15.8}
\newcommand{\nCostCampTaskMin}{12.5}

\newcommand{\nCostMarginTaskMax}{6.0}
\newcommand{\nCostMarginTaskMin}{4.4}

\newcommand{\nCtlAOneDReg}{\ensuremath{-}0.14}
\newcommand{\nCtlAOneDRegHi}{\ensuremath{-}0.03}
\newcommand{\nCtlAOneDRegLo}{\ensuremath{-}0.27}

\newcommand{\nCtlAallDReg}{\ensuremath{-}0.24}
\newcommand{\nCtlAallDRegHi}{\ensuremath{-}0.03}
\newcommand{\nCtlAallDRegLo}{\ensuremath{-}0.51}

\newcommand{\nCtlArDReg}{\ensuremath{-}0.09}
\newcommand{\nCtlArDRegHi}{\ensuremath{-}0.01}
\newcommand{\nCtlArDRegLo}{\ensuremath{-}0.18}

\newcommand{\nCtlCands}{74}

\newcommand{\nCtlProbeAOne}{1,040}

\newcommand{\nCtlProbeAall}{2,960}

\newcommand{\nCtlTasks}{37}

\newcommand{\nDdCands}{74}

\newcommand{\nDdFaster}{72}

\newcommand{\nDdSRInc}{97.5}
\newcommand{\nDdSRRep}{97.7}

\newcommand{\nDdStepsRel}{15.0}

\newcommand{\nDecisionPoints}{304}

\newcommand{\nDiagTasks}{72}

\newcommand{\nDnCands}{44}

\newcommand{\nDnRepairTasks}{22}

\newcommand{\nEpsFull}{2,550}
\newcommand{\nEpsRace}{1,186}

\newcommand{\nExDevMInc}{0.010}

\newcommand{\nExtStepsMax}{20.0}
\newcommand{\nExtStepsMin}{18.0}

\newcommand{\nFRFixedFive}{20}

\newcommand{\nFreshAUC}{0.774}
\newcommand{\nFreshAUCHi}{0.89}
\newcommand{\nFreshAUCLo}{0.63}

\newcommand{\nFreshFailHigh}{1.35}
\newcommand{\nFreshFailInLow}{18}
\newcommand{\nFreshFailLow}{4.00}
\newcommand{\nFreshFails}{41}
\newcommand{\nFreshLowSharePct}{21}

\newcommand{\nFreshRR}{2.97}
\newcommand{\nFreshRRHi}{7.85}
\newcommand{\nFreshRRLo}{1.03}

\newcommand{\nFreshTasks}{72}
\newcommand{\nFreshWsAUC}{0.689}

\newcommand{\nGoldHeldInc}{1.33}

\newcommand{\nGoldHeldRep}{0.07}

\newcommand{\nGoldReproduced}{193}

\newcommand{\nHeldFreshRho}{0.77}
\newcommand{\nHeldoutFailInSat}{53}

\newcommand{\nHeldoutFailTotal}{73}
\newcommand{\nLnAUC}{0.697}
\newcommand{\nLnAUCHi}{0.811}
\newcommand{\nLnAUCLo}{0.567}

\newcommand{\nLnDevPerfect}{90}

\newcommand{\nLnOneSteps}{18.0}

\newcommand{\nLnRR}{3.67}

\newcommand{\nLnRaceSave}{57.2}
\newcommand{\nLnRate}{99.2}
\newcommand{\nLnTasks}{90}

\newcommand{\nLnTwoSRRep}{99.6}
\newcommand{\nLnTwoSteps}{18.0}

\newcommand{\nMatrixTasks}{74}
\newcommand{\nNonAblCands}{170}

\newcommand{\nOffRate}{98.75}

\newcommand{\nPerfectDecisionPct}{69}

\newcommand{\nQOneHiFresh}{0/30}

\newcommand{\nQOneLoFresh}{5/30}
\newcommand{\nQOneLoSlack}{10}
\newcommand{\nQThreeGain}{200}

\newcommand{\nReproN}{4,000}
\newcommand{\nReproPass}{3,927}

\newcommand{\nReproRate}{98.175}
\newcommand{\nReproTasks}{80}
\newcommand{\nRoEpsA}{876}
\newcommand{\nRoEpsB}{414}

\newcommand{\nRoHeldA}{477}
\newcommand{\nRoHeldB}{469}

\newcommand{\nRoUSDA}{114}
\newcommand{\nRoUSDB}{112}
\newcommand{\nRoVersA}{59}
\newcommand{\nRoVersB}{28}

\newcommand{\nRsFails}{5}

\newcommand{\nRsMinMarginFails}{4}

\newcommand{\nRxAUC}{0.750}
\newcommand{\nRxAUCHi}{1.000}
\newcommand{\nRxAUCLo}{0.200}

\newcommand{\nRxDevPerfect}{6}

\newcommand{\nRxOneSteps}{20.0}

\newcommand{\nRxRaceSave}{57.1}
\newcommand{\nRxRate}{99.3}
\newcommand{\nRxTasks}{7}

\newcommand{\nRxTwoSRRep}{99.6}
\newcommand{\nRxTwoSteps}{18.4}

\newcommand{\nSatTasks}{78}
\newcommand{\nSatTasksMatrix}{72}

\newcommand{\nSaveRace}{53.5}
\newcommand{\nSaveRaceVsSoP}{15.0}

\newcommand{\nSkillRefuted}{3}
\newcommand{\nSkillVerified}{72}

\newcommand{\nTbBudgetInc}{3.3}
\newcommand{\nTbBudgetRep}{0.8}
\newcommand{\nTbCands}{30}

\newcommand{\nTbPNinetyInc}{891}
\newcommand{\nTbPNinetyRep}{788}
\newcommand{\nTbRec}{34}
\newcommand{\nTbRecFromBudget}{25}

\newcommand{\nTbSRInc}{94.0}
\newcommand{\nTbSRRep}{94.8}

\newcommand{\nTbStepsRel}{10.7}

\newcommand{\nTbTasks}{15}

\newcommand{\nUnionTasks}{37}

\usepackage{algorithm}
\usepackage{algpseudocode}
\usepackage{float} 
\definecolor{cvprblue}{rgb}{0.21,0.49,0.74}
\usepackage[pagebackref,breaklinks,colorlinks,allcolors=cvprblue]{hyperref}

\def\confName{CVPR}
\def\confYear{2027}

\title{MARGIN: A Risk-Aware Agent for Robot Program Refinement}
\author{
{\small Hongbo Duan$^{1,2}$\hspace{0.95em}Fan Du$^{3}$\hspace{0.95em}Haipeng Luo$^{1}$\hspace{0.95em}Bin Qian$^{1}$\hspace{0.95em}Zichao Nie$^{1}$\hspace{0.95em}Yilin Mi$^{1}$}\\[0.25em]
{\small Yaopeng Tian$^{1}$\hspace{0.95em}Jiake Tian$^{2}$\hspace{0.95em}Siyuan Liu$^{4}$\hspace{0.95em}Yan Gao$^{2}$\hspace{0.95em}Fangming Liu$^{2}$\hspace{0.95em}Xueqian Wang$^{1}$}\\[0.55em]
{\footnotesize $^{1}$Tsinghua University\quad $^{2}$Peng Cheng Laboratory}\\[0.12em]
{\footnotesize $^{3}$Southern University of Science and Technology\quad $^{4}$Institute of Automation, Chinese Academy of Science}
}

\begin{document}
\raggedbottom 
\maketitle
\vspace{-8pt}
\begin{strip}
\centering
\includegraphics[width=\linewidth]{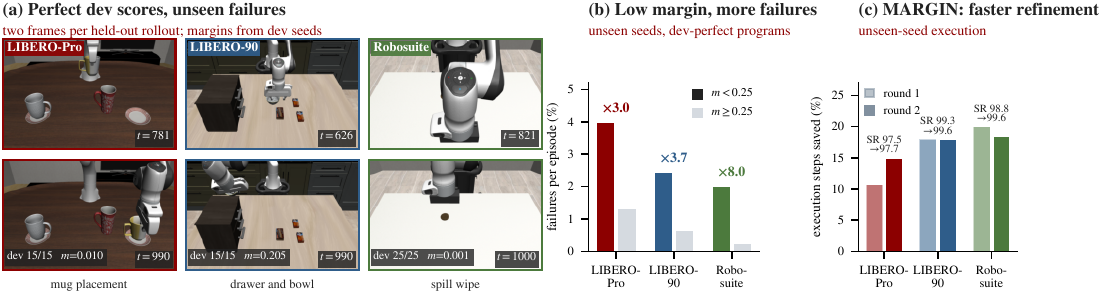}
\captionsetup{hypcap=false}
\captionof{figure}{\textbf{Success is not enough, on three benchmarks.}
Agentic debugging drives nearly every program to a perfect development score, yet
these programs can fail on unseen scenes.
\textbf{(a)}~Two real frames from the same failed unseen-seed rollout of a dev-perfect
program on each benchmark: LIBERO-Pro (mug placement), LIBERO-90 (drawer and bowl),
and Robosuite (spill wipe). The labels show the development success and the margin $m$
computed from development rollouts.
\textbf{(b)}~Dev-perfect programs with $m<0.25$ exhibit higher failure rates on unseen
scenes (LIBERO-Pro: fresh seeds; LIBERO-90 and Robosuite: held-out seeds).
\textbf{(c)}~\ours refinement reduces execution steps on unseen seeds across both rounds.}
\label{fig:teaser}
\end{strip}

\begin{abstract}
Robot-programming agents improve executable control policies through repeated testing and repair, yet binary success saturates while execution risk persists across unseen scenes. We introduce \ours, a risk-aware agent that converts successful robot rollouts into actionable evidence for program refinement. Its \emph{near-miss margin} measures the fraction of the execution budget remaining in the slowest successful development rollout. The agent uses this signal to prioritize fragile programs, guide targeted repairs, and select revisions that preserve development success while increasing execution headroom. An exact safe-racing rule accelerates success-based candidate evaluation. We evaluate \ours across LIBERO-Pro, LIBERO-90, and Robosuite. In our 80-task LIBERO-Pro reproduction, \nSatTasks{} programs attain perfect development success while accounting for substantial failures on unseen seeds. The near-miss margin predicts held-out failures with an AUC of \nFreshAUC, and low-margin programs exhibit \nFreshRR$\times$ higher failure risk. Across two independent evaluation rounds spanning \nUnionTasks{} unique LIBERO-Pro tasks, \ours reduces executed control steps by \nTbStepsRel\% and \nDdStepsRel\%, with reductions of \nExtStepsMin--\nExtStepsMax\% on LIBERO-90 and Robosuite. Safe racing reduces fixed-sequence candidate-evaluation episodes by \nSaveRace\% while retaining identical selections, and paired execution probes improve acceptance decisions by identifying hidden regressions. Together, these results establish near-miss evidence as an effective basis for risk-aware, budget-efficient robot program refinement.
\end{abstract}

\section{Introduction}
\label{sec:intro}
\begin{figure*}[t]
\centering
\includegraphics[width=\linewidth]{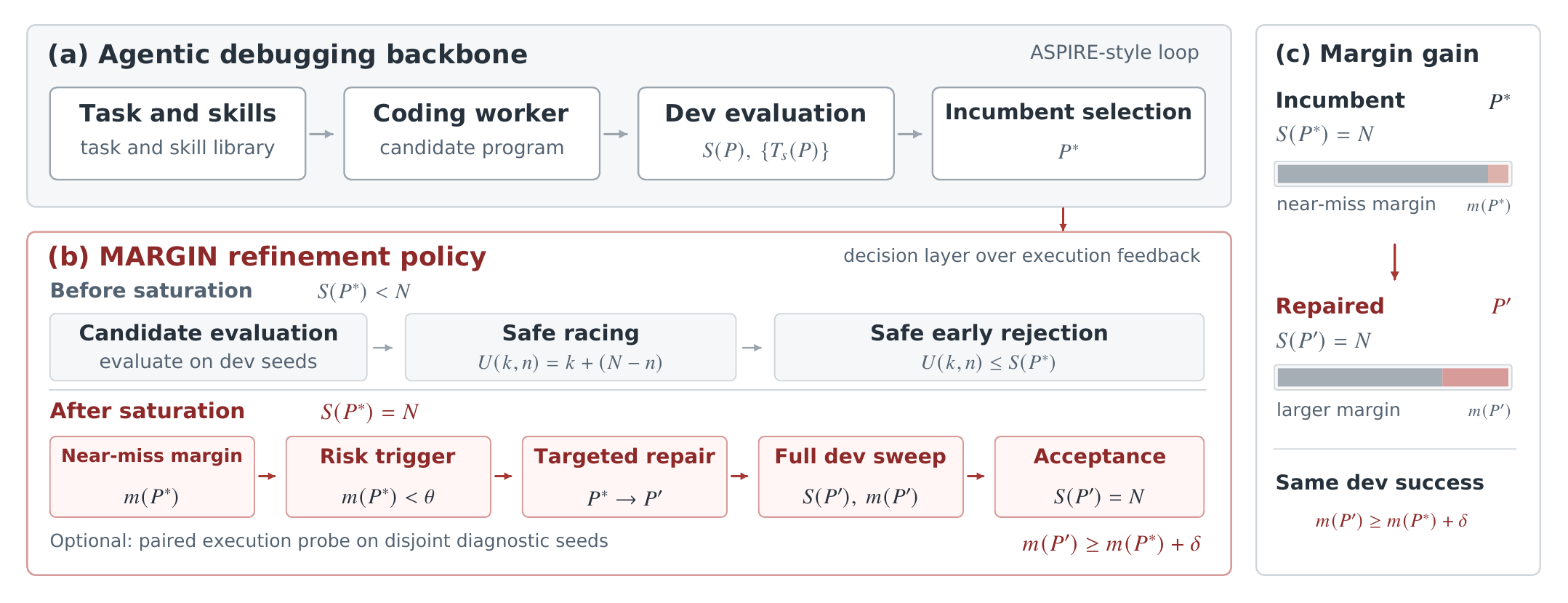}
\caption{\textbf{MARGIN: risk-aware refinement within agentic robot debugging.}
\textbf{(a)}~The debugging backbone combines task context, an LLM coding worker, development evaluation,
and incumbent selection; each rollout yields a success indicator and execution length.
\textbf{(b)}~Before saturation, safe racing evaluates candidates under strict success-count improvement,
rejecting a candidate when its upper bound $U(k,n)$ cannot exceed $S(P^\ast)$.
After $S(P^\ast)=N$, successful development rollouts provide the near-miss margin $m(P^\ast)$.
The risk trigger allocates repair budget $B$; a complete development sweep recomputes $S(P')$ and
$m(P')$, and the controller accepts only success-preserving, margin-improving audited revisions.
Optional paired probes screen hidden regressions on separate diagnostic seeds.
\textbf{(c)}~An illustrative margin increase with unchanged development success.}
\label{fig:method}
\end{figure*}

Language-model agents increasingly translate task instructions into executable robot programs, using perception, planning, and feedback to debug behavior across physical scenes~\cite{liang2023code,singh2023progprompt,huang2023voxposer,wang2023voyager,fu2026capx,lu2026aspire,wen2026roborsi,ge2026urai,gao2026hexashell}. \aspire~\cite{lu2026aspire} provides a representative agentic debugging loop: a coding worker composes perception, grasp synthesis, and motion-planning tools, evaluates candidate programs on development scenes, and consolidates successful revisions into a shared skill library. Our reproduction spans all \nReproTasks{} LIBERO-Pro tasks~\cite{zhou2025liberopro} and achieves \nReproRate\% success over \nReproN{} held-out episodes. This setting exposes a new opportunity for improvement: once development success saturates, execution traces still distinguish programs by how much room remains for reliable completion.

Consider two programs that each solve every development scene. One completes its most demanding rollout with substantial execution budget remaining; the other finishes with only \nQOneLoSlack{} steps left under the $\Tmax$-step limit (\cref{fig:teaser}a). On fresh scenes, the two programs incur \nQOneHiFresh{} and \nQOneLoFresh{} failures, respectively, despite identical development scores. This pattern appears at scale: \nSatTasks{} of the \nReproTasks{} selected programs achieve perfect development success, \nPerfectDecisionPct\% of candidate decisions face a perfect incumbent, and these programs still account for \nHeldoutFailInSat{} of \nHeldoutFailTotal{} held-out failures. We refer to this regime as \emph{success saturation}. It motivates a question central to agentic program refinement: \emph{how can evidence from successful executions identify remaining risk and guide the next refinement decision?}

We introduce \textbf{\ours} (Margin-Aware Refinement Guided by Informative Near-misses), a risk-aware agent that extends agentic debugging beyond the binary success criterion (\cref{fig:method}). Its \emph{near-miss margin} measures the residual execution budget in the slowest successful development rollout, providing a continuous signal from trials already performed. MARGIN prioritizes development-perfect programs with low margins and instructs the coding worker to improve their execution through generalizable program edits. A complete development sweep evaluates each revision, and a constrained acceptance rule selects programs that retain success while increasing the margin. During success-based search, an exact safe-racing rule reduces candidate-evaluation episodes without changing the strict-success selection criterion. The two mechanisms connect efficient program discovery with targeted refinement after saturation.

Our contributions are threefold:
\begin{itemize}
\item \textbf{Near-miss risk estimation.} We characterize success saturation in agentic robot programming and establish the near-miss margin as a risk signal for development-perfect programs. On fresh LIBERO-Pro seeds, it attains AUC \nFreshAUC{} and a \nFreshRR$\times$ low-margin risk ratio; the analysis extends to LIBERO-90, where AUC reaches \nLnAUC{} and the risk ratio reaches \nLnRR, as well as Robosuite (\cref{sec:exp-risk}).
\item \textbf{Risk-guided program refinement.} MARGIN couples margin-based repair prioritization with success-preserving acceptance, reducing unseen-scene execution steps by \nTbStepsRel\% and \nDdStepsRel\% across two LIBERO-Pro rounds and by \nExtStepsMin--\nExtStepsMax\% on LIBERO-90 and Robosuite (\cref{sec:exp-refine}).
\item \textbf{Budget-efficient evaluation and execution evidence.} Exact safe racing reduces fixed-sequence evaluation episodes by \nSaveRace\% with identical candidate selections, with corresponding savings of \nLnRaceSave\% and \nRxRaceSave\% on LIBERO-90 and Robosuite. Paired execution probes support more informed acceptance by detecting hidden regressions (\cref{sec:exp-cost,sec:exp-evidence}).
\end{itemize}

\section{Related Work}
\label{sec:related}

\noindent\textbf{Robot-programming agents and executable control.}
Language-model-driven robot systems combine semantic planning with executable policies and physical feedback. Code as Policies, ProgPrompt, VoxPoser, and Inner Monologue ground language instructions in robot programs, spatial representations, and execution observations~\cite{liang2023code,singh2023progprompt,huang2023voxposer,huang2022inner}. \aspire develops robot programs through iterative debugging and consolidates validated behaviors into reusable skills~\cite{lu2026aspire}. PyRUA-Lean, RobotUse, URAI, HexaShell, and uni-robot-agent expose robot capabilities through programmable interfaces and agent-controlled tool execution~\cite{si2026pyrualean,lee2026robotuse,ge2026urai,gao2026hexashell,yuan2026unirobot}. Embodied Turing Machines and agentic world-action models connect executable programs to persistent state and predictive control~\cite{hu2026etm,liu2026agenticwam}, while CaP-X, Robo-Harness K1, and CaPtain study code generation and robot-policy integration~\cite{fu2026capx,li2026k1,fan2026captain}. Vision-language-action models provide complementary visuomotor capabilities~\cite{kim2024openvla,black2024pi0,li2026simplevlarl}.

\noindent\textbf{Experience-driven refinement and recovery.}
Embodied agents increasingly use execution experience to revise programs, skills, and the tools through which they act. Voyager and Eureka establish iterative skill and reward-program improvement~\cite{wang2023voyager,ma2024eureka}; RoboRSI, PhysEvo, ENPIRE, and EMHO extend this direction to robot self-improvement, tool revision, and agent-harness optimization~\cite{wen2026roborsi,tian2026physevo,xiao2026enpire,lee2026emho}. Self-Refine, Reflexion, self-debugging, and CodeAct develop general mechanisms for learning from feedback through language and executable actions~\cite{madaan2023selfrefine,shinn2023reflexion,chen2024selfdebug,wang2024codeact}. REFLECT, AHA, DoReMi, Sentinel, BT-Agent, and REMM contribute multimodal error diagnosis, execution monitoring, and structured recovery~\cite{liu2023reflect,duan2024aha,guo2024doremi,agia2024sentinel,hubel2026btagent,senthilkumaran2026remm}. These systems establish execution traces as a foundation for adaptive robot behavior; \ours exploits the temporal information in \emph{successful} traces to guide refinement after binary success saturation.

\noindent\textbf{Robustness measurement and evaluation efficiency.}
Temporal-logic robustness quantifies the degree to which trajectories satisfy task specifications~\cite{fainekos2009robustness,donze2010robust}. LIBERO-PRO, LIBERO-Pro2.0, LIBERO-VPro, and SIMPLER test robotic behavior under scene, task, and observation variations~\cite{zhou2025liberopro,zhou2026liberopro2,li2026liberovpro,li2024simpler}; RobotWorld, Embodied Agent Arena, and LIBERO-Agent extend evaluation to robot-use agents and execution-level failure analysis~\cite{yang2026robotworld,huang2026arena,diao2026liberoagent}. Racing, successive halving, and active evaluation reduce the resources required to compare candidate systems~\cite{maron1994hoeffding,jamieson2016nonstochastic,li2018hyperband,anwar2025active,snyder2026beyond}. \ours combines exact success-count racing with a near-miss margin that directs refinement effort toward development-perfect programs with limited execution headroom.

\section{Method}
\label{sec:method}
\begin{figure*}[t]
\centering
\includegraphics[width=\linewidth]{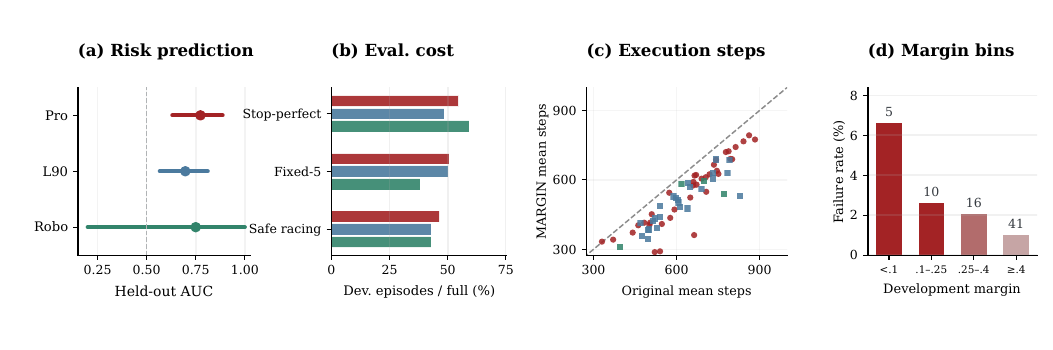}
\caption{\textbf{MARGIN results across three benchmarks.} Red, blue, and green indicate LIBERO-Pro, LIBERO-90, and Robosuite. \textbf{(a)} Task-level near-miss risk prediction with 95\% bootstrap intervals. \textbf{(b)} Development evaluation cost relative to full candidate sweeps, with Safe Racing preserving candidate selections in fixed-sequence replay. \textbf{(c)} Paired original and refined execution steps on unseen initializations. \textbf{(d)} Fresh-seed failure rate across LIBERO-Pro margin bins; values above bars show task counts.}
\label{fig:main}
\end{figure*}

\subsection{Setting: agentic program debugging}
\label{sec:setting}
Each task $\tau$ specifies a manipulation objective and a simulator that samples an initial scene from seed $s$.
The coding agent produces a program $P$ over perception, grasp-synthesis, and motion-planning
tools~\cite{lu2026aspire,sundaralingam2023curobo,sundermeyer2021contact}. Execution produces a
binary outcome $y_s(P)\in\{0,1\}$ and control-step count $T_s(P)\le\Tmax$, where $\Tmax=1000$.
In the underlying debugging loop, the worker proposes $P'$ and evaluates it on $N=15$ development
seeds $\Sdev$. The incumbent $P^\ast$ is updated whenever
$S(P')=\sum_{s\in\Sdev} y_s(P')$ exceeds the current development score. We study the
corresponding failure risk over unseen initializations,
\begin{equation}
  R(P)=\Pr_{s\sim p_{\mathrm{test}}}\bigl[y_s(P)=0\bigr],
  \label{eq:risk}
\end{equation}
with $S(P)$ serving as its development-set estimator.

\paragraph{Success saturation.}
When $S(P^\ast)=N$, the development success count has reached its maximum. In our campaign,
\nPerfectDecisionPct\% of the \nDecisionPoints{} candidate decisions face a perfect incumbent,
and \nSatTasksMatrix{} of the \nMatrixTasks{} tasks with complete candidate histories reach this
regime (\cref{fig:teaser}). Successful trajectories then offer a complementary signal for
refinement beyond the saturated score.

\subsection{Near-miss margin}
\label{sec:margin}
We quantify the execution headroom contained in successful development rollouts. For a program
with at least one successful episode, its \emph{near-miss margin} is
\begin{equation}
  m(P) \;=\; \min_{s\in\Sdev:\,y_s(P)=1}\Bigl(1-\frac{T_s(P)}{\Tmax}\Bigr)\in[0,1],
  \label{eq:margin}
\end{equation}
the fraction of the step budget remaining in the slowest successful episode. For development-perfect
programs, the minimum spans all development seeds. In our running example,
$m=\nExDevMInc$ because the slowest successful rollout has only \nQOneLoSlack{} steps
remaining. Retries, re-grasps, and motion detours reduce this margin and expose long-tail execution
cost. Taking the minimum emphasizes demanding scenes rather than averaging their effects. The
signal is computed directly from recorded executions, without additional trials.

\subsection{Risk-guided refinement budget}
\label{sec:budget}
MARGIN continues refinement after a task reaches maximum development success. A task enters the
refinement pool according to its incumbent margin:
\begin{equation}
  \text{refine}(\tau) \;=\; \bigl[S(P^\ast)=N\bigr]\wedge\bigl[m(P^\ast)<\theta\bigr],
  \label{eq:budget}
\end{equation}
We use $\theta=0.25$ in the first round and extend the eligible margin range to $m<0.5$ in
the second, following the respective evaluation protocols.

\subsection{Margin-aware repair and constrained acceptance}
\label{sec:accept}
For each selected task, the coding worker receives the incumbent, its development rollouts, and
the skill library. It proposes generalizable edits that reduce execution steps in demanding scenes,
such as consolidating motion segments, removing redundant waypoints, and simplifying wrist
rotations (\cref{fig:qual}a). The worker operates under the original program-provenance rules and
an explicit per-session budget of 105 development episodes, together with wall-clock and LLM-cost
limits. A seed-restricted execution wrapper enforces the development partition.

The controller evaluates proposed repairs on a complete development sweep. A candidate $P'$ is
eligible to replace $P^\ast$ when
\begin{equation}
  S(P')=N \;\wedge\; m(P')\ge m(P^\ast)+\delta \;\wedge\; \text{audit}(P'),
  \label{eq:accept}
\end{equation}
where $\delta=0.02$ corresponds to 20 control steps, and $\text{audit}$ checks program
provenance and infrastructure integrity. Among eligible candidates, the controller retains the
version with the largest margin. This policy couples complete development success with greater
execution headroom.

\subsection{Budget-efficient evaluation by safe racing}
\label{sec:racing}
During success-based debugging, a candidate must exceed the incumbent development score
$S^\ast=S(P^\ast)$. A full comparison evaluates $N$ seeds. After $n$ evaluations with $k$
successes, the maximum attainable score is
\begin{equation}
  U(k,n)=k+(N-n),\qquad\text{reject as soon as } U(k,n)\le S^\ast .
  \label{eq:race}
\end{equation}
The bound preserves strict-improvement selections while avoiding the remaining evaluations of
candidates that cannot win. The seed order prioritizes scenes on which earlier versions failed.
Margin-based repair is evaluated separately with complete development sweeps.

\begin{algorithm}[!t]
\small
\caption{\ours: Margin-aware refinement on top of an agentic debugging loop}
\label{alg:margin}
\begin{algorithmic}[1]
\Require Tasks $\mathcal T$; incumbents $P_\tau^\ast$; dev seeds $\Sdev$; thresholds $\theta,\delta$; budget $B$; replicates $R$
\For{each task $\tau\in\mathcal T$}
  \State $S^\ast\gets S(P_\tau^\ast)$; $m^\ast\gets m(P_\tau^\ast)$
  \If{$S^\ast<N$}
    \State Continue ordinary debugging with safe racing
  \ElsIf{$m^\ast<\theta$} \Comment{dev-perfect near miss}
    \For{$r=1,\ldots,R$}
      \State $\mathcal C_r\gets\mathrm{Repair}(P_\tau^\ast,\Sdev,B)$
      \State Keep $P'\in\mathcal C_r$ only if $S(P')=N$,
      \Statex \hspace{2.1em}$m(P')\ge m^\ast+\delta$, and audit passes
      \State Select eligible $P'$ with largest $m(P')$;
      \Statex \hspace{2.1em}otherwise keep $P_\tau^\ast$
      \State Optionally reject upon net probe regressions
    \EndFor
  \EndIf
\EndFor
\end{algorithmic}
\end{algorithm}

\subsection{Execution probes}
\label{sec:probe}
For additional evidence, the controller can execute the candidate and incumbent on $k$ paired
\emph{diagnostic} seeds, disjoint from development and final evaluation. It compares regressions
(incumbent succeeds, candidate fails) against recoveries (candidate succeeds, incumbent fails) and
retains the incumbent when regressions dominate. \Cref{sec:exp-evidence} studies this execution
probe alongside trace-derived alarms and keyframe-based VLM diagnoses.

\section{Experiments}
\label{sec:exp}
\begin{table*}[!t]
\centering
\caption{\textbf{LIBERO-Pro success rates across eight task categories.}
P/T: position/task perturbations. Bold: best in each comparable column (ties included).
The six-category mean is followed by the eight-category mean in parentheses when available. Only our ASPIRE reproduction is listed.
MARGIN combines ASPIRE incumbents and accepted refinements over 80 tasks.
Baseline scores are from the cited papers.}
\label{tab:lit-success}
\begingroup
\fontsize{8.2}{9.7}\selectfont
\setlength{\tabcolsep}{3.8pt}
\renewcommand{\arraystretch}{1.03}
\begin{tabular*}{\textwidth}{@{\extracolsep{\fill}}l*{9}{r}@{}}
\toprule
& \multicolumn{2}{c}{Object} & \multicolumn{2}{c}{Spatial} & \multicolumn{2}{c}{Goal} & \multicolumn{2}{c}{LIBERO-10} & \multicolumn{1}{c}{Mean (\%)} \\
\cmidrule(lr){2-3}\cmidrule(lr){4-5}\cmidrule(lr){6-7}\cmidrule(lr){8-9}\cmidrule(l){10-10}
Method & P & T & P & T & P & T & P & T & \multicolumn{1}{c}{Avg.6 (Avg.8)} \\
\midrule
\multicolumn{10}{@{}l}{\emph{Policy}} \\
$\pi_{0.5}$~\cite{fu2026capx} & 17.0 & 1.0 & 20.0 & 1.0 & 38.0 & 0.0 & -- & -- & 12.8 \\
\addlinespace[0.6pt]
\multicolumn{10}{@{}l}{\emph{Agent + policy}} \\
Pigey~\cite{galanti2026pigey} & 54.0 & 54.0 & 66.0 & 80.0 & 44.0 & 22.0 & -- & -- & 53.3 \\
Harness VLA (Codex)~\cite{zhang2026harnessvla} & 91.0 & 94.0 & 69.0 & 81.0 & 66.0 & 75.0 & 49.0 & 52.0 & 79.3 (72.1) \\
Harness VLA (CC)~\cite{zhang2026harnessvla} & 90.0 & 88.0 & 80.0 & 94.0 & 87.0 & 87.0 & 62.0 & 71.0 & 87.7 (82.4) \\
\addlinespace[0.6pt]
\multicolumn{10}{@{}l}{\emph{Robot-use agent}} \\
CaP-X~\cite{fu2026capx} & 22.0 & 18.0 & 12.0 & 14.0 & 26.0 & 17.0 & -- & -- & 18.2 \\
RHO~\cite{elmaaroufi2026rho} & 12.2 & 37.2 & 53.0 & 61.6 & 50.6 & 55.6 & -- & -- & 45.0 \\
RATS~\cite{zhang2026rats} & 61.0 & 63.0 & 29.0 & 31.0 & 43.0 & 36.0 & -- & -- & 43.8 \\
OpenRUA~\cite{chu2026openrua} & 98.0 & 97.0 & 98.0 & \textbf{100.0} & 79.0 & 79.0 & 80.0 & 65.0 & 91.8 (87.0) \\
HexaShell + skills~\cite{gao2026hexashell} & 96.0 & \textbf{100.0} & 97.0 & 96.2 & \textbf{99.0} & \textbf{100.0} & \textbf{94.0} & 96.7 & 98.0 (97.4) \\
\midrule
ASPIRE (Opus 5.5)$^{\dagger}$~\cite{lu2026aspire} & \textbf{99.2} & \textbf{100.0} & \textbf{98.8} & 98.4 & 98.4 & 98.8 & 93.8 & \textbf{98.0} & 98.9 (\textbf{98.2}) \\
\rowcolor{oursbg}\textbf{MARGIN} & \textbf{99.2} & \textbf{100.0} & 98.6 & 98.9 & 98.5 & 99.2 & 93.4 & 97.8 & \textbf{99.1} (\textbf{98.2}) \\
\bottomrule
\end{tabular*}
\vspace{2pt}
\parbox{\textwidth}{\fontsize{7.5}{8.5}\selectfont
$^{\dagger}$Our reproduction of ASPIRE with Opus 5.5.}
\endgroup
\end{table*}

\subsection{Setup}
\label{sec:exp-setup}
\paragraph{Benchmarks.}
We evaluate MARGIN on 80 LIBERO-Pro tasks~\cite{zhou2025liberopro}, 90 LIBERO-90 tasks~\cite{liu2023libero}, and seven Robosuite tasks~\cite{zhu2020robosuite}. LIBERO-Pro covers the Object, Spatial, Goal, and LIBERO-10 suites under position and task perturbations. We reproduce \aspire~\cite{lu2026aspire} with Claude Opus 5.5 and its official skill-library workflow. Each task uses seeded scenes and a 1000-step horizon.

\paragraph{Protocol.}
On LIBERO-Pro, development seeds 51--65 guide candidate selection; held-out seeds 1--50 and independent fresh seeds 101--130 evaluate risk. Refinement uses seeds 131--180 and 231--280, with diagnostic probes on 181--200. We pre-register evaluation conditions and freeze decisions before final testing. The task is the resampling unit for 95\% confidence intervals, computed from 2000 bootstrap samples. Each evaluation uses a frozen program version and task-specific seed partition. The controller consumes only development evidence; unseen-scene results are collected after selection.

\paragraph{Metrics.}
We measure task success, mean and P90 control steps, development episodes, and agent cost. Programs are evaluated on paired initial states with recorded program hashes, rollout lengths, and final task outcomes. High-step failures terminate after at least 950 control steps. Independent repair sessions share the same incumbent and frozen skill library.

\begin{figure*}[t]
\centering
\includegraphics[width=\linewidth]{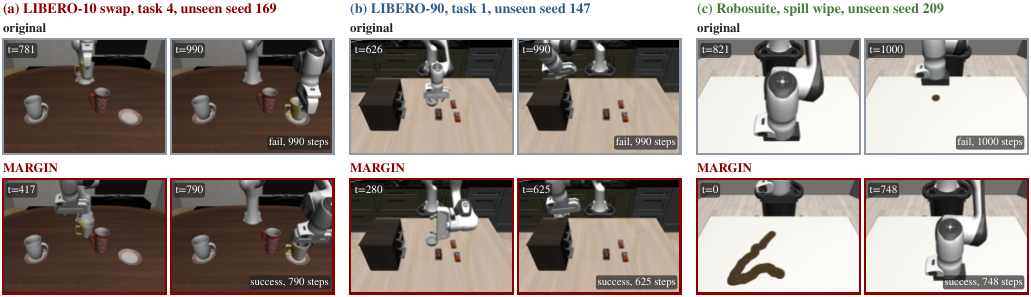}
\caption{\textbf{MARGIN refinement examples across three benchmarks.} Real simulator frames illustrate paired execution outcomes on unseen initializations. Top: the original program at matched control steps and termination. Bottom: the MARGIN revision during execution and at termination. The variable $t$ denotes the executed control step. \textbf{(a)} LIBERO-Pro mug placement. The revision completes placement with \nQThreeGain{} fewer steps. \textbf{(b)} LIBERO-90 drawer closure and bowl placement. The revision completes the task with ample execution headroom. \textbf{(c)} Robosuite spill wiping. The revision completes the wipe in a shorter trajectory.}
\label{fig:qual}
\end{figure*}

\subsection{Success-rate comparisons and success saturation}
\label{sec:exp-repro}
\begin{table*}[!t]
\centering

\begin{minipage}[t]{0.485\textwidth}
\centering
\caption{\textbf{LIBERO-90 success rates.} Baseline scores are from the cited papers.}
\label{tab:lit-libero90}
\begingroup\fontsize{8.2}{9.5}\selectfont\setlength{\tabcolsep}{6pt}
\begin{tabular*}{\linewidth}{@{\extracolsep{\fill}}lr@{}}
\toprule Method & Success (\%) \\
\midrule
DiffuView~\cite{zhang2026diffuview} & 92.5 \\
TRM-VLA~\cite{li2026trmvla} & 94.8 \\
AtlasVLA~\cite{zhao2026atlasvla} & 95.8 \\
LaMem-VLA~\cite{qu2026lamemvla} & 97.0 \\
GeoVLA~\cite{sun2025geovla} & 97.7 \\
MINT-4B~\cite{huang2026mint} & 98.7 \\
\midrule
ASPIRE (Opus 5.5)$^{\dagger}$~\cite{lu2026aspire} & 99.2 \\
\rowcolor{oursbg} \textbf{MARGIN} & \textbf{99.6} \\
\bottomrule\end{tabular*}\endgroup
\end{minipage}
\hfill
\begin{minipage}[t]{0.485\textwidth}
\centering
\caption{\textbf{Robosuite manipulation success rates.} Baseline scores are from the cited papers.}
\label{tab:lit-robosuite}
\begingroup\fontsize{8.2}{9.5}\selectfont\setlength{\tabcolsep}{6pt}
\begin{tabular*}{\linewidth}{@{\extracolsep{\fill}}lr@{}}
\toprule Method & Success (\%) \\
\midrule
CaP-X~\cite{fu2026capx} & 68.3 \\
RHO~\cite{elmaaroufi2026rho} & 70.0 \\
Robo-Harness K1~\cite{li2026k1} & 72.1 \\
HarnessPAI~\cite{wang2026harnesspai} & 96.3 \\
OpenRUA~\cite{chu2026openrua} & 99.0 \\
\midrule
ASPIRE (Opus 5.5)$^{\dagger}$~\cite{lu2026aspire} & 99.3 \\
\rowcolor{oursbg} \textbf{MARGIN} & \textbf{99.6} \\
\bottomrule\end{tabular*}\endgroup
\end{minipage}

\end{table*}

\Cref{tab:lit-success,tab:lit-libero90,tab:lit-robosuite} compares published methods and our ASPIRE--MARGIN evaluations. Across the full 80-task LIBERO-Pro suite, MARGIN achieves 98.2\% success. MARGIN reaches 99.6\% success in the LIBERO-90 and Robosuite refinement evaluations, alongside the execution-step gains reported in \cref{tab:refine}.

\textbf{Success saturation motivates further refinement.} Our ASPIRE reproduction succeeds in \nReproPass/\nReproN{} LIBERO-Pro held-out episodes for a success rate of \nReproRate\% and reaches perfect development success on \nSatTasks{} tasks. Of their \nHeldoutFailTotal{} held-out failures, \nHeldoutFailInSat{} occur on dev-perfect tasks. On LIBERO-90 and Robosuite, ASPIRE reaches \nLnRate\% and \nRxRate\% held-out success, with \nLnDevPerfect/\nLnTasks{} and \nRxDevPerfect/\nRxTasks{} dev-perfect programs. The official Opus 5.5 ASPIRE evaluation reaches \nOffRate\% on the 80 LIBERO-Pro tasks, while our reproduced programs achieve \nReproRate\%. The skill library contains \nSkillVerified{} verified and \nSkillRefuted{} refuted entries. The fine-grained results in \cref{tab:lit-success} reveal strong saturation in the Object suites and concentrated failures under LIBERO-10 perturbations, motivating margin-aware refinement beyond binary success.

\subsection{Near-miss margin predicts residual risk}
\label{sec:exp-risk}
\begin{table}[t]
\centering\small
\setlength{\tabcolsep}{3.8pt}
\caption{\textbf{Near-miss risk prediction across benchmarks.} Task-level failure prediction for dev-perfect programs, evaluated on unseen seeds.}
\label{tab:risk-main}
\begin{tabular*}{\columnwidth}{@{\extracolsep{\fill}}lccc@{}}
\toprule
Benchmark & Programs & AUC [95\% CI] & Risk ratio \\
\midrule
LIBERO-Pro & \nFreshTasks{} & \nFreshAUC{} [\nFreshAUCLo,\,\nFreshAUCHi] & \nFreshRR \\
LIBERO-90 & \nLnDevPerfect{} & \nLnAUC{} [\nLnAUCLo,\,\nLnAUCHi] & \nLnRR \\
Robosuite & \nRxDevPerfect{} & \nRxAUC{} [\nRxAUCLo,\,\nRxAUCHi] & -- \\
\bottomrule
\end{tabular*}
\end{table}

\textbf{The near-miss margin predicts held-out risk.} On \nDiagTasks{} dev-perfect programs, fresh-seed evaluation yields AUC \nFreshAUC{} [\nFreshAUCLo, \nFreshAUCHi]. The low-margin group fails on \nFreshFailLow\% of unseen episodes versus \nFreshFailHigh\% for the remaining programs, a risk ratio of \nFreshRR{} [\nFreshRRLo, \nFreshRRHi], as shown in \cref{fig:main}d. Within-suite AUC reaches \nFreshWsAUC. LIBERO-90 replicates the signal with AUC \nLnAUC{} [\nLnAUCLo, \nLnAUCHi] and risk ratio \nLnRR. In Robosuite, the lowest-margin program accounts for \nRsMinMarginFails{} of \nRsFails{} held-out failures in \cref{fig:main}a. Across the LIBERO-Pro tasks, the low-margin subset constitutes \nFreshLowSharePct\% of programs but explains \nFreshFailInLow{} of \nFreshFails{} failures on fresh seeds. Held-out and fresh-seed failure counts show Spearman correlation \nHeldFreshRho, giving a consistent ranking across distinct evaluation rounds. 

\subsection{Margin-aware refinement}
\label{sec:exp-refine}
\begin{table*}[t]
\centering\small
\setlength{\tabcolsep}{4pt}
\caption{\textbf{Margin-aware refinement on unseen seeds.} Two independent repairs per task. Round 2 includes 37 unique tasks, including 22 newly selected programs. Best value per Original/MARGIN pair in bold (higher SR, lower steps/failure rates). Bootstrap intervals for reduction and failure change.}
\label{tab:refine}
\begin{adjustbox}{max width=\textwidth}
\begin{tabular}{llcccccccccc}
\toprule
Eval.\ seeds & Tasks & Program & SR (\%) & Mean & Median & P90 & Worst & High-step (\%) & Incomplete (\%) & Step red.\ (\%) & $\Delta$Fail (pp) \\
\midrule
R1: 131--180 & 15 & Original & 94.0 & 776 & 763 & 891 & 902 & 3.3 & \textbf{2.7} & & \\
\rowcolor{oursbg} & 30 prog. & MARGIN & \textbf{94.8} & \textbf{693} & \textbf{690} & \textbf{788} & \textbf{809} & \textbf{0.8} & 4.4 & 10.7 [9.3, 12.2] & $-$0.80 [$-$3.27, +0.87] \\
\midrule
R2: 231--280 & 37 & Original & 97.5 & 648 & 665 & 824 & 779 & 0.8 & \textbf{1.7} & & \\
\rowcolor{oursbg} & 74 prog. & MARGIN & \textbf{97.7} & \textbf{550} & \textbf{578} & \textbf{740} & \textbf{662} & \textbf{0.1} & 2.1 & 15.4 [12.4, 19.1] & $-$0.22 [$-$0.92, +0.30] \\
\midrule
R2, new tasks only & 22 & Original & \textbf{98.4} & 562 & 573 & 696 & 701 & 0.4 & \textbf{1.3} & & \\
\rowcolor{oursbg} & 44 prog. & MARGIN & 98.3 & \textbf{455} & \textbf{440} & \textbf{608} & \textbf{573} & \textbf{0.1} & 1.6 & 18.5 [13.9, 23.6] & +0.05 [$-$0.91, +0.86] \\
\bottomrule
\end{tabular}
\end{adjustbox}
\end{table*}

\begin{figure}[t]
\centering
\includegraphics[width=\columnwidth]{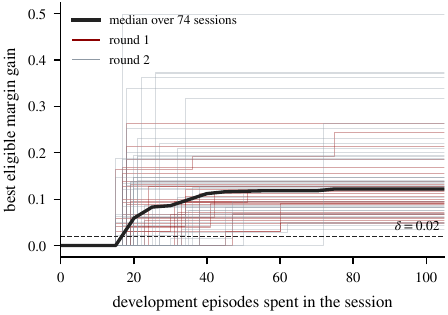}
\caption{\textbf{Margin gains during agentic repair.} Each curve traces the best eligible margin improvement within a refinement session as more development episodes are evaluated. The solid black curve summarizes the median across 74 repair sessions; the dashed line marks the acceptance threshold $\delta=0.02$.}
\label{fig:margin-trajectories}
\end{figure}

MARGIN performs two independent repair sessions per task. Round~1 refines \nTbTasks{} low-margin LIBERO-Pro programs; round~2 adds \nDnRepairTasks{} tasks under the expanded threshold, for \nUnionTasks{} unique tasks. The \nTbCands{} and \nDnCands{} repair sessions satisfy the acceptance rule in \cref{eq:accept}. Their margin improvements accumulate across development evaluations, as illustrated in \cref{fig:margin-trajectories}.

\begin{figure}[H]
\centering
\includegraphics[width=\columnwidth]{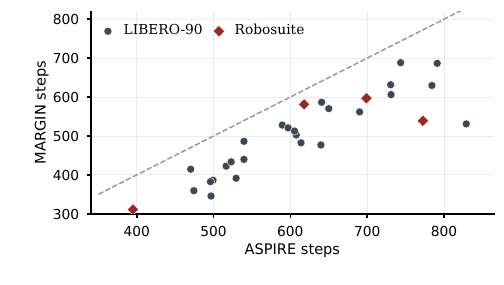}
\caption{\textbf{Cross-benchmark repair gains.} Per-task mean control steps for ASPIRE and MARGIN on independent LIBERO-90 and Robosuite seeds. Each point averages two repair runs; the dashed line denotes equal steps.}
\label{fig:transfer-steps}
\end{figure}

\textbf{Repairs shorten unseen-scene execution.} Round~1 reduces mean control steps by \nTbStepsRel\%, with P90 steps decreasing from \nTbPNinetyInc{} to \nTbPNinetyRep{}, and round~2 by \nDdStepsRel\%; \nDdFaster{} of \nDdCands{} programs are faster. High-step failures decrease from \nTbBudgetInc\% to \nTbBudgetRep\%. Held-out success changes from \nTbSRInc\% to \nTbSRRep\% in round~1 and from \nDdSRInc\% to \nDdSRRep\% in round~2. On LIBERO-90 and Robosuite, the two rounds shorten execution by \nLnOneSteps--\nLnTwoSteps\% and \nRxOneSteps--\nRxTwoSteps\%, respectively; success reaches \nLnTwoSRRep\% and \nRxTwoSRRep\% as shown in \cref{tab:refine,fig:main}c. Round 2 extends refinement to intermediate-margin programs. Each session evaluates every development seed, retaining only candidates that preserve development success and improve the minimum remaining-step margin. Task-level and pooled results quantify the gain on unseen initial states.
\textbf{Per-task transfer.} Across 25 LIBERO-90 tasks and four Robosuite tasks in the second refinement round, all 58 independently accepted repairs reduce mean control steps on unseen scenes, as shown in \cref{fig:transfer-steps}. The LIBERO-90 programs decrease from 613 to 503 control steps on average, with eight recoveries from step-limit failures. Robosuite gains span wiping, assembly, stacking, and bimanual manipulation.

\begin{table}[t]
\centering\small
\setlength{\tabcolsep}{4pt}
\caption{\textbf{Agent cost by benchmark.} USD per task for \aspire discovery and \ours refinement; total USD for refinement.}
\label{tab:taskcost-main}
\begin{tabular*}{\columnwidth}{@{\extracolsep{\fill}}lrrrr@{}}
\toprule
Benchmark & Tasks & \aspire & \ours & Total \\
\midrule
LIBERO-Pro & 37 & -- & 4.4 & 164 \\
LIBERO-90 & 25 & 12.5 & 6.0 & 151 \\
Robosuite & 4 & 15.8 & 5.8 & 23 \\
\midrule
Total & 66 & -- & 5.1 & 338 \\
\bottomrule
\end{tabular*}
\end{table}

\textbf{Refinement cost and recovery mechanisms.} Repairing a program costs \$\nCostMarginTaskMin--\nCostMarginTaskMax{} per task, compared with \$\nCostCampTaskMin--\nCostCampTaskMax{} for the ASPIRE discovery campaign in \cref{tab:taskcost-main}. Replaying \nGoldReproduced{} episodes with an analysis-only evaluator shows that \nTbRecFromBudget{} of \nTbRec{} recoveries originate from high-step failures; the fraction ending with an object still grasped falls from \nGoldHeldInc\% to \nGoldHeldRep\% in \cref{fig:qual}. The paired trajectories in \cref{fig:qual} illustrate the reduction in control steps and recoveries from near-budget executions. Every first-round repair and \nDdFaster{} of \nDdCands{} second-round repairs shorten execution. Paired tests record recovery and regression events on the same initial scenes.

\subsection{Budget-efficient evaluation}
\label{sec:exp-cost}
\begin{table}[t]
\centering\small
\setlength{\tabcolsep}{4.5pt}
\caption{\textbf{Evaluation efficiency.} Offline replay on 170 non-ablation candidates (fixed sequence); live 10-task agent experiment (different search trajectories).}
\label{tab:eval-main}
\begin{tabular*}{\columnwidth}{@{\extracolsep{\fill}}lrrr@{}}
\toprule
Offline selection & Dev ep. & Saved (\%) & False rej. \\
\midrule
Full-15 & 2,550 & -- & \textbf{0} \\
Stop-on-Perfect & 1,395 & 45.3 & \textbf{0} \\
Fixed-5 strict & 1,290 & 49.4 & 20 \\
\rowcolor{oursbg}Safe racing & \textbf{1,186} & \textbf{53.5} & \textbf{0} \\
\midrule
Live agent (10 tasks) & Dev ep. & Held-out & LLM USD \\
\midrule
Full-sweep agent & 876 & \textbf{477/500} & 113.5 \\
\rowcolor{oursbg}Safe-racing agent & \textbf{414} & 469/500 & \textbf{112.0} \\
\bottomrule
\end{tabular*}
\end{table}

\textbf{Safe racing reduces development evaluation.} On a fixed sequence of \nNonAblCands{} non-ablation candidates, safe racing evaluates \nEpsRace{} rather than \nEpsFull{} episodes, saving \nSaveRace\%, with the same selected programs. It reduces evaluation by \nSaveRaceVsSoP\% relative to Stop-on-Perfect. Fixed-5 testing instead rejects \nFRFixedFive{} true improvements. On LIBERO-90 and Robosuite, racing saves \nLnRaceSave\% and \nRxRaceSave\% of episodes. In a 10-task live comparison, development episodes fall from \nRoEpsA{} to \nRoEpsB{}, while held-out successes are \nRoHeldA/500 and \nRoHeldB/500 in \cref{tab:eval-main,fig:main}b. The live agents generate \nRoVersA{} and \nRoVersB{} candidate versions, respectively, with LLM costs of \$\nRoUSDA{} and \$\nRoUSDB. Fixed-sequence replay preserves selected programs, while online agents adapt subsequent candidate generation to the evaluation feedback.

\begin{figure*}[t]
\centering
\includegraphics[width=\linewidth]{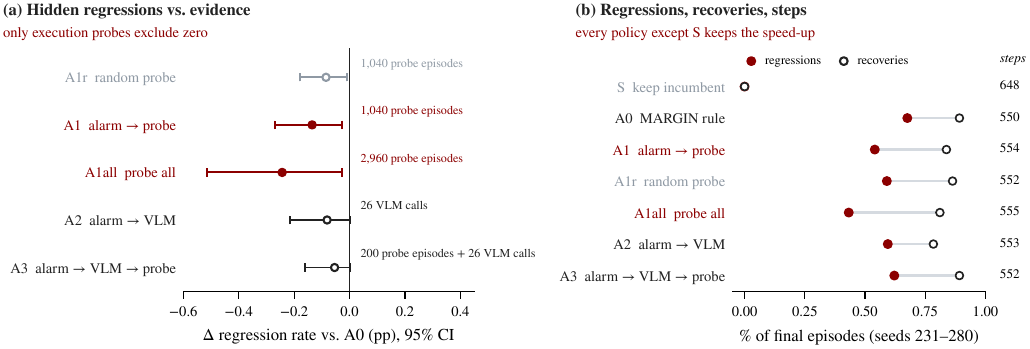}
\caption{\textbf{Execution probes detect hidden regressions.} The study covers \nCtlCands{} candidates and uses decisions frozen before evaluation on independent seeds. \textbf{(a)} Change in regression rate relative to accepting every MARGIN candidate, together with the evidence cost of each policy. \textbf{(b)} Regression and recovery rates alongside execution steps under each policy.}
\label{fig:cost}
\end{figure*}

\subsection{Which evidence catches hidden regressions?}
\label{sec:exp-evidence}
\begin{table}[t]
\centering\small
\setlength{\tabcolsep}{4.3pt}
\caption{\textbf{Evidence-policy ablation.} Round 2, 37 tasks and 74 candidate programs; final-seed regression and recovery rates (\%).}
\label{tab:evidence-main}
\begin{tabular*}{\columnwidth}{@{\extracolsep{\fill}}lrrrr@{}}
\toprule
Policy & Probe ep. & VLM & Regr. & Recov. \\
\midrule
A0: MARGIN & 0 & 0 & 0.68 & \textbf{0.89} \\
\rowcolor{oursbg}A1: alarm $\to$ probe & 1,040 & 0 & 0.54 & 0.84 \\
A1r: random probe & 1,040 & 0 & 0.59 & 0.86 \\
A1all: probe all & 2,960 & 0 & \textbf{0.43} & 0.81 \\
A2: VLM verdict & 0 & 26 & 0.59 & 0.78 \\
A3: VLM + probe & 200 & 26 & 0.62 & \textbf{0.89} \\
\bottomrule
\end{tabular*}
\end{table}

We compare seven acceptance policies on \nCtlCands{} candidate programs across \nCtlTasks{} tasks, using trace alarms, paired executions, and VLM diagnoses (\cref{tab:evidence-main,fig:cost}).

\textbf{Execution probes reduce hidden regressions.} Alarm-triggered probing (A1) reduces the regression rate by \nCtlAOneDReg{} pp [\nCtlAOneDRegLo, \nCtlAOneDRegHi]; probing every candidate (A1all) reduces it by \nCtlAallDReg{} pp [\nCtlAallDRegLo, \nCtlAallDRegHi]. Cost-matched random probing (A1r) achieves a comparable reduction of \nCtlArDReg{} pp [\nCtlArDRegLo, \nCtlArDRegHi]. These policies preserve execution-step gains, while VLM-only filtering (A2) and VLM-assisted probing (A3) produce smaller changes. The full-probe policy consumes \nCtlProbeAall{} additional episodes, compared with \nCtlProbeAOne{} for alarm-triggered testing. Independent paired seeds measure regressions and recoveries, with cost-matched random probing separating evidence quantity from selection strategy.

\textbf{Trace-guided probe selection.} The alarm rule compares each candidate with its incumbent on identical development seeds. It monitors changes in gripper aperture, end-effector release height, failed tool calls, and gripper-command counts. An alarm triggers an execution check on 20 paired diagnostic seeds, yielding both recovery events, where the revision succeeds after incumbent failure, and regression events, where the revision fails after incumbent success. The decision retains a candidate when the number of recoveries is at least the number of regressions. This gives the agent a task-outcome criterion for selecting revisions alongside the margin-based development acceptance rule. The paired cases in \cref{fig:qual} exhibit both recovered completions and incomplete placements, linking execution evidence to the final acceptance decision. Equivalently, the net diagnostic outcome is
$D(P',P^\ast)=\sum_{s\in\mathcal S_{\mathrm{probe}}}[y_s(P')-y_s(P^\ast)]$:
a recovery contributes $+1$, a regression $-1$, and an agreement zero.
The decision requires $D\geq0$ independently of the margin gain, so development-time
execution efficiency and diagnostic success preservation remain separate tests.

\FloatBarrier
\section{Conclusion}
\label{sec:conclusion}
We introduced \ours, a risk-aware robot-programming agent that uses near-miss evidence from successful executions to guide refinement after binary success saturates. Its temporal margin identifies programs with limited execution headroom, and constrained acceptance turns this signal into targeted code improvements. Across LIBERO-Pro, LIBERO-90, and Robosuite, MARGIN shortens execution by \nTbStepsRel--\nExtStepsMax\% on unseen initial states. Safe racing reduces success-based candidate evaluation, while paired execution probes expose hidden regressions before program adoption. Together, these mechanisms connect program generation, efficiency-oriented refinement, and execution-grounded verification in one agentic loop. The results establish successful trajectories as decision evidence beyond task completion and motivate richer interaction-aware margins.

\par\medskip
\section*{LLM Usage Statement}
Claude Opus 5.5 powers the robot-programming agents and the visual critic. Claude Code was used for
implementation, experiment orchestration, statistical analysis, figure and \LaTeX{} scripts, and
manuscript drafting. ChatGPT assisted with manuscript revision, literature analysis, and figure
review. The authors verified the experimental results, citations, and final manuscript.

{
    \small
    \bibliographystyle{ieeenat_fullname}
    \bibliography{main}

\begin{thebibliography}{64}
\providecommand{\natexlab}[1]{#1}
\providecommand{\url}[1]{\texttt{#1}}
\expandafter\ifx\csname urlstyle\endcsname\relax
  \providecommand{\doi}[1]{doi: #1}\else
  \providecommand{\doi}{doi: \begingroup \urlstyle{rm}\Url}\fi

\bibitem[Agia et~al.(2024)Agia, Sinha, Yang, Cao, Antonova, Pavone, and
  Bohg]{agia2024sentinel}
Christopher Agia, Rohan Sinha, Jingyun Yang, Zi-ang Cao, Rika Antonova, Marco
  Pavone, and Jeannette Bohg.
\newblock Unpacking failure modes of generative policies: Runtime monitoring of
  consistency and progress.
\newblock In \emph{Conference on Robot Learning (CoRL)}, 2024.

\bibitem[Anwar et~al.(2025)Anwar, Gupta, Merchant, Ghosh, Neiswanger, and
  Thomason]{anwar2025active}
Abrar Anwar, Rohan Gupta, Zain Merchant, Sayan Ghosh, Willie Neiswanger, and
  Jesse Thomason.
\newblock Efficient evaluation of multi-task robot policies with active
  experiment selection.
\newblock \emph{arXiv preprint arXiv:2502.09829}, 2025.

\bibitem[Black et~al.(2024)Black, Brown, Driess, Esmail, Equi, Finn, Fusai,
  Groom, Hausman, Ichter, Jakubczak, Jones, Ke, Levine, Li-Bell, Mothukuri,
  Nair, Pertsch, Shi, Tanner, Vuong, Walling, Wang, and
  Zhilinsky]{black2024pi0}
Kevin Black, Noah Brown, Danny Driess, Adnan Esmail, Michael Equi, Chelsea
  Finn, Niccolo Fusai, Lachy Groom, Karol Hausman, Brian Ichter, Szymon
  Jakubczak, Tim Jones, Liyiming Ke, Sergey Levine, Adrian Li-Bell, Mohith
  Mothukuri, Suraj Nair, Karl Pertsch, Lucy~Xiaoyang Shi, James Tanner, Quan
  Vuong, Anna Walling, Haohuan Wang, and Ury Zhilinsky.
\newblock $\pi_0$: A vision-language-action flow model for general robot
  control.
\newblock \emph{arXiv preprint arXiv:2410.24164}, 2024.

\bibitem[Chen et~al.(2024)Chen, Lin, Sch{\"a}rli, and Zhou]{chen2024selfdebug}
Xinyun Chen, Maxwell Lin, Nathanael Sch{\"a}rli, and Denny Zhou.
\newblock Teaching large language models to self-debug.
\newblock In \emph{International Conference on Learning Representations
  (ICLR)}, 2024.

\bibitem[Chu et~al.(2026)Chu, Barr, Le~Goues, O'Hearn, Harman, Sarro, and
  Ye]{chu2026openrua}
Zhaoyang Chu, Earl~T. Barr, Claire Le~Goues, Peter O'Hearn, Mark Harman,
  Federica Sarro, and He Ye.
\newblock {OpenRUA}: Robot-use agents are zero-shot visuomotor policies.
\newblock \emph{arXiv preprint arXiv:2610.02459}, 2026.

\bibitem[Diao et~al.(2026)Diao, Chen, Xie, Lu, Peng, Zhong, Xu, Ye, Wu, and
  Jiang]{diao2026liberoagent}
Zijie Diao, Yitong Chen, Sicheng Xie, Tianyi Lu, Wujian Peng, Guojin Zhong,
  Houze Xu, Ziyi Ye, Zuxuan Wu, and Yu-Gang Jiang.
\newblock {LIBERO-Agent: Evaluating General-Purpose Agents for Direct Embodied
  Manipulation}.
\newblock \emph{arXiv preprint arXiv:2609.39507}, 2026.

\bibitem[Donz{\'e} and Maler(2010)]{donze2010robust}
Alexandre Donz{\'e} and Oded Maler.
\newblock Robust satisfaction of temporal logic over real-valued signals.
\newblock In \emph{International Conference on Formal Modeling and Analysis of
  Timed Systems (FORMATS)}, pages 92--106, 2010.

\bibitem[Duan et~al.(2024)Duan, Pumacay, Kumar, Wang, Tian, Yuan, Krishna, Fox,
  Mandlekar, and Guo]{duan2024aha}
Jiafei Duan, Wilbert Pumacay, Nishanth Kumar, Yi~Ru Wang, Shulin Tian, Wentao
  Yuan, Ranjay Krishna, Dieter Fox, Ajay Mandlekar, and Yijie Guo.
\newblock {AHA}: A vision-language-model for detecting and reasoning over
  failures in robotic manipulation.
\newblock \emph{arXiv preprint arXiv:2410.00371}, 2024.

\bibitem[Elmaaroufi et~al.(2026)Elmaaroufi, Svegliato, Kalade, Schelle, Seshia,
  and Zaharia]{elmaaroufi2026rho}
Karim Elmaaroufi, Justin Svegliato, Sarunas Kalade, Graham Schelle, Sanjit~A.
  Seshia, and Matei Zaharia.
\newblock {RHO}: Your coding agent is secretly a roboticist.
\newblock \emph{arXiv preprint arXiv:2606.16458}, 2026.

\bibitem[Fainekos and Pappas(2009)]{fainekos2009robustness}
Georgios~E. Fainekos and George~J. Pappas.
\newblock Robustness of temporal logic specifications for continuous-time
  signals.
\newblock \emph{Theoretical Computer Science}, 410\penalty0 (42):\penalty0
  4262--4291, 2009.

\bibitem[Fan et~al.(2026)Fan, He, Jiang, Wu, Wang, and Yue]{fan2026captain}
Yijia Fan, Xilin He, Yilei Jiang, Dongming Wu, Jiaxu Wang, and Xiangyu Yue.
\newblock {CaPtain}: A pipeline for post-training code-as-policy robot agents.
\newblock \emph{arXiv preprint arXiv:2610.40799}, 2026.

\bibitem[Fu et~al.(2026)Fu, Yu, El-Refai, Kou, Xue, Huang, Xiao, Wang, Niu,
  Fei-Fei, Shi, Wu, Sastry, Zhu, Goldberg, and Fan]{fu2026capx}
Letian Fu, Justin Yu, Karim El-Refai, Ethan Kou, Haoru Xue, Huang Huang, Wenli
  Xiao, Guanzhi Wang, Dantong Niu, Li Fei-Fei, Guanya Shi, Jiajun Wu, Shankar
  Sastry, Yuke Zhu, Ken Goldberg, and Linxi Fan.
\newblock {CaP-X}: A framework for benchmarking and improving coding agents for
  robot manipulation.
\newblock \emph{arXiv preprint arXiv:2603.22435}, 2026.

\bibitem[Galanti et~al.(2026)Galanti, Shah, and Dao]{galanti2026pigey}
Liane Galanti, Dhruv Shah, and Tri Dao.
\newblock Addressing the orchestration gap in generalist robots via physical
  agency.
\newblock \emph{arXiv preprint arXiv:2607.21725}, 2026.

\bibitem[Gao et~al.(2026)Gao, Zeng, Zhou, Ge, Shangguan, Wang, Wan, Zheng, Zhu,
  Li, Wu, Wang, He, Jia, Wu, and He]{gao2026hexashell}
Hongcheng Gao, Jianshu Zeng, Alex Zhou, Shijia Ge, Xuan Shangguan, Yazhe Wang,
  Yexing Wan, Zelin Zheng, Jay Zhu, Xiping Li, Di Wu, Yongli Wang, Lingyu He,
  Zhiqi Jia, Sihang Wu, and Xiao He.
\newblock {HexaShell: A Universal Harness for Robot Coding Agents in Simulation
  and the Real World}.
\newblock \emph{arXiv preprint arXiv:2610.23479}, 2026.

\bibitem[Ge et~al.(2026)Ge, Zhou, Zeng, Wan, Wu, Zheng, Wang, Jia, Shangguan,
  Zhu, Liu, He, Wu, He, and Gao]{ge2026urai}
Shijia Ge, Alex Zhou, Jianshu Zeng, Yexing Wan, Di Wu, Zelin Zheng, Yazhe Wang,
  Zhiqi Jia, Xuan Shangguan, Jay Zhu, Yijun Liu, Lingyu He, Sihang Wu, Xiao He,
  and Hongcheng Gao.
\newblock {Make Code as Policy Great Again: Frontier Agents Write, Call, and
  Evolve Robot Tools}.
\newblock \emph{arXiv preprint arXiv:2609.39018}, 2026.

\bibitem[Guo et~al.(2024)Guo, Wang, Zha, and Chen]{guo2024doremi}
Yanjiang Guo, Yen-Jen Wang, Lihan Zha, and Jianyu Chen.
\newblock {DoReMi: Grounding Language Model by Detecting and Recovering from
  Plan-Execution Misalignment}.
\newblock In \emph{IEEE/RSJ International Conference on Intelligent Robots and
  Systems (IROS)}, 2024.

\bibitem[Hu et~al.(2026)Hu, Hu, Hong, Chen, and Liu]{hu2026etm}
Kairui Hu, Siyuan Hu, Fangzhou Hong, Zhaoxi Chen, and Ziwei Liu.
\newblock {Embodied Turing Machines: Stateful Code for Robot Recursive
  Self-Improvement}.
\newblock \emph{arXiv preprint arXiv:2610.12369}, 2026.

\bibitem[Huang et~al.(2026{\natexlab{a}})Huang, Zhao, Li, Zhang, Wei, Li, Yang,
  Zhou, Chen, and Li]{huang2026arena}
Haojian Huang, Pukun Zhao, Zexi Li, Yehang Zhang, Yangkai Wei, Wenqian Li, Han
  Yang, Kaiwen Zhou, Ying-Cong Chen, and Yinchuan Li.
\newblock {Are Frontier VLM Agents Ready to Be Robot Generalists? An Empirical
  Study with the Embodied Agent Arena}.
\newblock \emph{arXiv preprint arXiv:2610.00854}, 2026{\natexlab{a}}.

\bibitem[Huang et~al.(2026{\natexlab{b}})Huang, Zeng, Tang, Cai, Lu, and
  Cai]{huang2026mint}
Renming Huang, Chendong Zeng, Wenjing Tang, Jingtian Cai, Cewu Lu, and Panpan
  Cai.
\newblock Mimic intent, not just trajectories.
\newblock \emph{arXiv preprint arXiv:2602.08602}, 2026{\natexlab{b}}.

\bibitem[Huang et~al.(2022)Huang, Xia, Xiao, Chan, Liang, Florence, Zeng,
  Tompson, Mordatch, Chebotar, Sermanet, Brown, Jackson, Luu, Levine, Hausman,
  and Ichter]{huang2022inner}
Wenlong Huang, Fei Xia, Ted Xiao, Harris Chan, Jacky Liang, Pete Florence, Andy
  Zeng, Jonathan Tompson, Igor Mordatch, Yevgen Chebotar, Pierre Sermanet, Noah
  Brown, Tomas Jackson, Linda Luu, Sergey Levine, Karol Hausman, and Brian
  Ichter.
\newblock Inner monologue: Embodied reasoning through planning with language
  models.
\newblock \emph{arXiv preprint arXiv:2207.05608}, 2022.

\bibitem[Huang et~al.(2023)Huang, Wang, Zhang, Li, Wu, and
  Fei-Fei]{huang2023voxposer}
Wenlong Huang, Chen Wang, Ruohan Zhang, Yunzhu Li, Jiajun Wu, and Li Fei-Fei.
\newblock {VoxPoser}: Composable 3d value maps for robotic manipulation with
  language models.
\newblock In \emph{Conference on Robot Learning (CoRL)}, 2023.

\bibitem[H{\"u}bel et~al.(2026)H{\"u}bel, Lilja, Yin, He, Tomlin, Sastry, Frey,
  and Pavone]{hubel2026btagent}
Fabio H{\"u}bel, Adam Lilja, Yue Yin, Siming He, Claire Tomlin, Shankar Sastry,
  Jonas Frey, and Marco Pavone.
\newblock {BT-Agent: Deploying Onboard Robot Agents with Behavior Trees and
  Multi-Fidelity Evaluation}.
\newblock \emph{arXiv preprint arXiv:2610.13862}, 2026.

\bibitem[Jamieson and Talwalkar(2016)]{jamieson2016nonstochastic}
Kevin Jamieson and Ameet Talwalkar.
\newblock Non-stochastic best arm identification and hyperparameter
  optimization.
\newblock In \emph{International Conference on Artificial Intelligence and
  Statistics (AISTATS)}, 2016.

\bibitem[Kim et~al.(2024)Kim, Pertsch, Karamcheti, Xiao, Balakrishna, Nair,
  Rafailov, Foster, Lam, Sanketi, Vuong, Kollar, Burchfiel, Tedrake, Sadigh,
  Levine, Liang, and Finn]{kim2024openvla}
Moo~Jin Kim, Karl Pertsch, Siddharth Karamcheti, Ted Xiao, Ashwin Balakrishna,
  Suraj Nair, Rafael Rafailov, Ethan Foster, Grace Lam, Pannag Sanketi, Quan
  Vuong, Thomas Kollar, Benjamin Burchfiel, Russ Tedrake, Dorsa Sadigh, Sergey
  Levine, Percy Liang, and Chelsea Finn.
\newblock {OpenVLA}: An open-source vision-language-action model.
\newblock \emph{arXiv preprint arXiv:2406.09246}, 2024.

\bibitem[Lee et~al.(2026{\natexlab{a}})Lee, Kim, Jeong, Kam, Kim, and
  Lee]{lee2026emho}
Hyun~Jung Lee, Jungtaek Kim, Jongwon Jeong, Tae-Eui Kam, Donghyun Kim, and
  Yong~Jae Lee.
\newblock {EMHO: EMbodied Agent Harness Optimization via Experience Traces}.
\newblock \emph{arXiv preprint arXiv:2610.08432}, 2026{\natexlab{a}}.

\bibitem[Lee et~al.(2026{\natexlab{b}})Lee, Baek, Kim, Jeon, Kim, Kim, and
  Kwak]{lee2026robotuse}
Junhoo Lee, Injun Baek, Seungyeon Kim, Suhyun Jeon, Minkyu Kim, Baekseung Kim,
  and Nojun Kwak.
\newblock {RobotUse: Allocating Computation, Context, and Decisions}.
\newblock \emph{arXiv preprint arXiv:2610.04929}, 2026{\natexlab{b}}.

\bibitem[Li et~al.(2026{\natexlab{a}})Li, Mei, Majumdar, Chen, Jiang, and
  Zhu]{li2026liberovpro}
Huiqiong Li, Zhiting Mei, Anirudha Majumdar, Jingjing Chen, Yu-Gang Jiang, and
  Bin Zhu.
\newblock {LIBERO-VPro}: Benchmarking closed-loop visual robustness of robotic
  foundation models.
\newblock \emph{arXiv preprint arXiv:2609.24350}, 2026{\natexlab{a}}.

\bibitem[Li et~al.(2026{\natexlab{b}})Li, Zuo, Yu, Zhang, Yang, Zhang, Zhu,
  Zhang, Chen, Cui, et~al.]{li2026simplevlarl}
Haozhan Li, Yuxin Zuo, Jiale Yu, Yuhao Zhang, Zhaohui Yang, Kaiyan Zhang,
  Xuekai Zhu, Yuchen Zhang, Tianxing Chen, Ganqu Cui, et~al.
\newblock {SimpleVLA-RL}: Scaling vla training via reinforcement learning.
\newblock In \emph{International Conference on Learning Representations
  (ICLR)}, 2026{\natexlab{b}}.

\bibitem[Li et~al.(2018)Li, Jamieson, DeSalvo, Rostamizadeh, and
  Talwalkar]{li2018hyperband}
Lisha Li, Kevin Jamieson, Giulia DeSalvo, Afshin Rostamizadeh, and Ameet
  Talwalkar.
\newblock Hyperband: A novel bandit-based approach to hyperparameter
  optimization.
\newblock \emph{Journal of Machine Learning Research}, 18\penalty0
  (185):\penalty0 1--52, 2018.

\bibitem[Li et~al.(2024)Li, Hsu, Gu, Pertsch, Mees, Walke, Fu, Lunawat, Sieh,
  Kirmani, Levine, Wu, Finn, Su, Vuong, and Xiao]{li2024simpler}
Xuanlin Li, Kyle Hsu, Jiayuan Gu, Karl Pertsch, Oier Mees, Homer~Rich Walke,
  Chuyuan Fu, Ishikaa Lunawat, Isabel Sieh, Sean Kirmani, Sergey Levine, Jiajun
  Wu, Chelsea Finn, Hao Su, Quan Vuong, and Ted Xiao.
\newblock Evaluating real-world robot manipulation policies in simulation.
\newblock In \emph{Conference on Robot Learning (CoRL)}, 2024.

\bibitem[Li et~al.(2026{\natexlab{c}})Li, Li, Wang, and Wang]{li2026trmvla}
Xiang Li, Ya-Li Li, Yuan Wang, and Shengjin Wang.
\newblock {TRM-VLA}: Temporal-aware chain-of-thought reasoning and memorization
  for vision-language-action models.
\newblock In \emph{IEEE/CVF Conference on Computer Vision and Pattern
  Recognition (CVPR)}, 2026{\natexlab{c}}.

\bibitem[Li et~al.(2026{\natexlab{d}})Li, Zhang, Li, Huang, Wang, Deng, Wei,
  Zhang, Xie, Zhou, et~al.]{li2026k1}
Zexi Li, Yehang Zhang, Wenqian Li, Haojian Huang, Chenxu Wang, Shiyuan Deng,
  Yangkai Wei, Tianyi Zhang, Binghui Xie, Bohan Zhou, et~al.
\newblock {Robo-Harness K1}: Harnessing robot-use agents via perception
  augmentation.
\newblock \emph{arXiv preprint arXiv:2609.29389}, 2026{\natexlab{d}}.

\bibitem[Liang et~al.(2023)Liang, Huang, Xia, Xu, Hausman, Ichter, Florence,
  and Zeng]{liang2023code}
Jacky Liang, Wenlong Huang, Fei Xia, Peng Xu, Karol Hausman, Brian Ichter, Pete
  Florence, and Andy Zeng.
\newblock Code as policies: Language model programs for embodied control.
\newblock In \emph{IEEE International Conference on Robotics and Automation
  (ICRA)}, pages 9493--9500, 2023.

\bibitem[Liu et~al.(2023{\natexlab{a}})Liu, Zhu, Gao, Feng, Liu, Zhu, and
  Stone]{liu2023libero}
Bo Liu, Yifeng Zhu, Chongkai Gao, Yihao Feng, Qiang Liu, Yuke Zhu, and Peter
  Stone.
\newblock {LIBERO}: Benchmarking knowledge transfer for lifelong robot
  learning.
\newblock \emph{arXiv preprint arXiv:2306.03310}, 2023{\natexlab{a}}.

\bibitem[Liu et~al.(2023{\natexlab{b}})Liu, Bahety, and Song]{liu2023reflect}
Zeyi Liu, Arpit Bahety, and Shuran Song.
\newblock {REFLECT}: Summarizing robot experiences for failure explanation and
  correction.
\newblock In \emph{Conference on Robot Learning (CoRL)}, 2023{\natexlab{b}}.

\bibitem[Liu et~al.(2026)Liu, Niu, Zhang, Shao, Wang, Zheng, Sharma, Lu, Liu,
  Zhang, Herzig, Bai, Goldberg, Xu, Zhu, Wang, Fan, and
  Darrell]{liu2026agenticwam}
Zhuoyang Liu, Dantong Niu, Botong Zhang, Boning Shao, Jing Wang, Ruijie Zheng,
  Yuvan Sharma, Runyu Lu, Jiaming Liu, Renrui Zhang, Roei Herzig, Yutong Bai,
  Ken Goldberg, Danfei Xu, Yuke Zhu, Guanzhi Wang, Linxi Fan, and Trevor
  Darrell.
\newblock {Agentic World Action Models}.
\newblock \emph{arXiv preprint arXiv:2610.30136}, 2026.

\bibitem[Lu et~al.(2026)Lu, Wu, Kou, Lin, Fu, Xiao, Mandlekar, Xu, Shi,
  Goldberg, Chen, Chowdhury, Kautz, Zhu, Fan, and Wang]{lu2026aspire}
Runyu Lu, Yubo Wu, Ethan Kou, Haotian Lin, Letian Fu, Wenli Xiao, Ajay
  Mandlekar, Yinzhen Xu, Guanya Shi, Ken Goldberg, Ang Chen, Mosharaf
  Chowdhury, Jan Kautz, Yuke Zhu, Linxi Fan, and Guanzhi Wang.
\newblock {ASPIRE}: Agentic /skills discovery for robotics.
\newblock \emph{arXiv preprint arXiv:2607.00272}, 2026.

\bibitem[Ma et~al.(2024)Ma, Liang, Wang, Huang, Bastani, Jayaraman, Zhu, Fan,
  and Anandkumar]{ma2024eureka}
Yecheng~Jason Ma, William Liang, Guanzhi Wang, De-An Huang, Osbert Bastani,
  Dinesh Jayaraman, Yuke Zhu, Linxi Fan, and Anima Anandkumar.
\newblock Eureka: Human-level reward design via coding large language models.
\newblock In \emph{International Conference on Learning Representations
  (ICLR)}, 2024.

\bibitem[Madaan et~al.(2023)Madaan, Tandon, Gupta, Hallinan, Gao, Wiegreffe,
  Alon, Dziri, Prabhumoye, Yang, Gupta, Majumder, Hermann, Welleck,
  Yazdanbakhsh, and Clark]{madaan2023selfrefine}
Aman Madaan, Niket Tandon, Prakhar Gupta, Skyler Hallinan, Luyu Gao, Sarah
  Wiegreffe, Uri Alon, Nouha Dziri, Shrimai Prabhumoye, Yiming Yang, Shashank
  Gupta, Bodhisattwa~Prasad Majumder, Katherine Hermann, Sean Welleck, Amir
  Yazdanbakhsh, and Peter Clark.
\newblock Self-refine: Iterative refinement with self-feedback.
\newblock In \emph{Advances in Neural Information Processing Systems
  (NeurIPS)}, 2023.

\bibitem[Maron and Moore(1993)]{maron1994hoeffding}
Oded Maron and Andrew~W. Moore.
\newblock Hoeffding races: Accelerating model selection search for
  classification and function approximation.
\newblock In \emph{Advances in Neural Information Processing Systems (NIPS)},
  1993.

\bibitem[Qu et~al.(2026)Qu, Gao, Hu, Yang, Yu, Yan, Wang, Shu, and
  Yan]{qu2026lamemvla}
Hongyu Qu, Jianzhe Gao, Xiaobin Hu, Shaohuan Yang, Xinlei Yu, Rui Yan, Wenguan
  Wang, Xiangbo Shu, and Shuicheng Yan.
\newblock Dual latent memory in vision-language-action models for robotic
  manipulation.
\newblock \emph{arXiv preprint arXiv:2607.07608}, 2026.

\bibitem[Senthilkumaran et~al.(2026)Senthilkumaran, Wickramarachchi, Oltramari,
  and Francis]{senthilkumaran2026remm}
Revanth~Krishna Senthilkumaran, Ruwan Wickramarachchi, Alessandro Oltramari,
  and Jonathan Francis.
\newblock {REMM: Revisable Embodied Memory for Manipulation}.
\newblock \emph{arXiv preprint arXiv:2610.12710}, 2026.

\bibitem[Shinn et~al.(2023)Shinn, Cassano, Berman, Gopinath, Narasimhan, and
  Yao]{shinn2023reflexion}
Noah Shinn, Federico Cassano, Edward Berman, Ashwin Gopinath, Karthik
  Narasimhan, and Shunyu Yao.
\newblock Reflexion: Language agents with verbal reinforcement learning.
\newblock In \emph{Advances in Neural Information Processing Systems
  (NeurIPS)}, 2023.

\bibitem[Si et~al.(2026)Si, Bi, Yang, Ni, Huang, Wang, Jiang, Zhang, Ruan,
  Zhang, Wang, Li, Feng, Dong, and Zhou]{si2026pyrualean}
Ruiyang Si, Jianxin Bi, Shunyu Yang, Rui Ni, Wenbo Huang, Qiang Wang, Shulong
  Jiang, Yintianrun Zhang, Feixiang Ruan, Kewei Zhang, Duomin Wang, Xiuyu Li,
  Haiwen Feng, Zhen Dong, and Daquan Zhou.
\newblock Fewer tokens, better action: {GPT-6 Astra} robot agents with 14\%
  higher success rate but 65\% fewer tokens.
\newblock \emph{arXiv preprint arXiv:2610.01939}, 2026.

\bibitem[Singh et~al.(2023)Singh, Blukis, Mousavian, Goyal, Xu, Tremblay, Fox,
  Thomason, and Garg]{singh2023progprompt}
Ishika Singh, Valts Blukis, Arsalan Mousavian, Ankit Goyal, Danfei Xu, Jonathan
  Tremblay, Dieter Fox, Jesse Thomason, and Animesh Garg.
\newblock {ProgPrompt}: Generating situated robot task plans using large
  language models.
\newblock In \emph{IEEE International Conference on Robotics and Automation
  (ICRA)}, pages 11523--11530, 2023.

\bibitem[Snyder et~al.(2026)Snyder, Badithela, Matni, Pappas, Majumdar, Itkina,
  and Nishimura]{snyder2026beyond}
David Snyder, Apurva Badithela, Nikolai Matni, George Pappas, Anirudha
  Majumdar, Masha Itkina, and Haruki Nishimura.
\newblock Beyond binary success: Sample-efficient and statistically rigorous
  robot policy comparison.
\newblock \emph{arXiv preprint arXiv:2603.13616}, 2026.

\bibitem[Sun et~al.(2025)Sun, Xie, Liu, Shi, Wang, and Cao]{sun2025geovla}
Lin Sun, Bin Xie, Yingfei Liu, Hao Shi, Tiancai Wang, and Jiale Cao.
\newblock {GeoVLA}: Empowering 3d representations in vision-language-action
  models.
\newblock \emph{arXiv preprint arXiv:2508.09071}, 2025.

\bibitem[Sundaralingam et~al.(2023)Sundaralingam, Hari, Fishman, Garrett,
  Van~Wyk, Blukis, Millane, Oleynikova, Handa, Ramos, Ratliff, and
  Fox]{sundaralingam2023curobo}
Balakumar Sundaralingam, Siva Kumar~Sastry Hari, Adam Fishman, Caelan Garrett,
  Karl Van~Wyk, Valts Blukis, Alexander Millane, Helen Oleynikova, Ankur Handa,
  Fabio Ramos, Nathan Ratliff, and Dieter Fox.
\newblock {cuRobo}: Parallelized collision-free minimum-jerk robot motion
  generation.
\newblock \emph{arXiv preprint arXiv:2310.17274}, 2023.

\bibitem[Sundermeyer et~al.(2021)Sundermeyer, Mousavian, Triebel, and
  Fox]{sundermeyer2021contact}
Martin Sundermeyer, Arsalan Mousavian, Rudolph Triebel, and Dieter Fox.
\newblock {Contact-GraspNet}: Efficient 6-dof grasp generation in cluttered
  scenes.
\newblock In \emph{IEEE International Conference on Robotics and Automation
  (ICRA)}, pages 13438--13444, 2021.

\bibitem[Tian et~al.(2026)Tian, Zhang, Liu, Liu, Liu, and
  Wang]{tian2026physevo}
Wenqing Tian, Zeyu Zhang, Zhaocheng Liu, Fengwei Liu, Qiang Liu, and Liang
  Wang.
\newblock {PhysEvo: Astra Can Act, Let It}.
\newblock \emph{arXiv preprint arXiv:2610.08995}, 2026.

\bibitem[Wang et~al.(2023)Wang, Xie, Jiang, Mandlekar, Xiao, Zhu, Fan, and
  Anandkumar]{wang2023voyager}
Guanzhi Wang, Yuqi Xie, Yunfan Jiang, Ajay Mandlekar, Chaowei Xiao, Yuke Zhu,
  Linxi Fan, and Anima Anandkumar.
\newblock Voyager: An open-ended embodied agent with large language models.
\newblock \emph{arXiv preprint arXiv:2305.16291}, 2023.

\bibitem[Wang et~al.(2024)Wang, Chen, Yuan, Zhang, Li, Peng, and
  Ji]{wang2024codeact}
Xingyao Wang, Yangyi Chen, Lifan Yuan, Yizhe Zhang, Yunzhu Li, Hao Peng, and
  Heng Ji.
\newblock Executable code actions elicit better {LLM} agents.
\newblock In \emph{International Conference on Machine Learning (ICML)}, 2024.

\bibitem[Wang et~al.(2026)Wang, Wu, Zhang, Wang, Shao, Luan, Li, Li, Gu,
  et~al.]{wang2026harnesspai}
Xin Wang, Wenhao Wu, Menghao Zhang, Zhi Wang, Kun Shao, Jian Luan, Yang Li,
  Qing Li, Shangding Gu, et~al.
\newblock {HarnessPAI}: An evolving harness for physical ai.
\newblock \emph{arXiv preprint arXiv:2609.29166}, 2026.

\bibitem[Wen et~al.(2026)Wen, Chen, Cao, Chen, Zou, Yu, Kuang, Xue, Lv, Wen,
  and Lu]{wen2026roborsi}
Zimo Wen, Yijin Chen, Yuxuan Cao, Wendi Chen, Yanwen Zou, Wenye Yu, Fuhang
  Kuang, Han Xue, Jun Lv, Chuan Wen, and Cewu Lu.
\newblock {RoboRSI: Stable, Efficient, and Reusable Robot Self-Evolution in
  Complex Real-World Environments}.
\newblock \emph{arXiv preprint arXiv:2610.12424}, 2026.

\bibitem[Xiao et~al.(2026)Xiao, Xie, Zhang, Lin, Fu, Xue, Lu, Yang, Dai, Wang,
  Wu, Wang, Sastry, Goldberg, Fan, Zhu, and Shi]{xiao2026enpire}
Wenli Xiao, Jia Xie, Tonghe Zhang, Haotian Lin, Letian Fu, Haoru Xue, Jalen Lu,
  Yi Yang, Cunxi Dai, Zi Wang, Jimmy Wu, Guanzhi Wang, S.~Shankar Sastry, Ken
  Goldberg, Linxi Fan, Yuke Zhu, and Guanya Shi.
\newblock {ENPIRE: Agentic Robot Policy Self-Improvement in the Real World}.
\newblock \emph{arXiv preprint arXiv:2606.19980}, 2026.

\bibitem[Yang et~al.(2026)Yang, Li, Hu, Liu, Huang, Sun, Li, Wu, Huang, Huang,
  Sun, Liu, Tong, Huang, Hu, Huang, Bai, Wang, Lin, Tang, Zhang, Chen, Song,
  Dai, Chen, Ji, Zhan, Hu, Xue, Zhang, Hu, Ho, and Guo]{yang2026robotworld}
Zhiqin Yang, Chenxin Li, Xiaomeng Hu, Yibin Liu, Weidong Huang, Jiankai Sun,
  Haitao Li, Zijian Wu, Yuzhi Huang, Fanding Huang, Hanwen Sun, Jiashun Liu,
  Jingqi Tong, Mingxin Huang, Shaoli Hu, Shijue Huang, Tianyi Bai, Xinyuan
  Wang, Yunlong Lin, Zhengyang Tang, Zhexin Zhang, Zhuo Chen, Xierui Song,
  Juntao Dai, Boyuan Chen, Jiaming Ji, Fangneng Zhan, Mengkang Hu, Wei Xue,
  Yonggang Zhang, Han Hu, Tsung-Yi Ho, and Yike Guo.
\newblock {RobotWorld: Benchmarking Multimodal Agents for Robot Use Across
  Diverse Tasks and Embodiments}.
\newblock \emph{arXiv preprint arXiv:2610.10409}, 2026.

\bibitem[Yuan et~al.(2026)Yuan, Zuo, Huang, Zhang, Park, Yang, Zhang, Chen,
  Ouyang, He, Lv, Wu, Liu, Zhao, Luo, Li, Qian, Ge, Gao, Yang, Deng, Xu, Huang,
  Zhou, Xu, Liu, Abbeel, and Ding]{yuan2026unirobot}
Jiarui Yuan, Yuxin Zuo, Xiaoyu Huang, Wenbo Zhang, Kwanyoung Park, Junlin Yang,
  Tianxiang Zhang, Ruoqu Chen, Yutao Ouyang, Bingxiang He, Xingtai Lv, Hongjin
  Wu, Jiajun Liu, Jiahao Zhao, Tianwei Luo, Haozhan Li, Cheng Qian, Jiaxin Ge,
  Huan-ang Gao, Kaisen Yang, Jiaqi Deng, Qixin Xu, Siqiao Huang, Bowen Zhou,
  Mengdi Xu, Zhiyuan Liu, Pieter Abbeel, and Ning Ding.
\newblock {Democratizing Robot-Use Agents: A Unified Harness with Open Models}.
\newblock \emph{arXiv preprint arXiv:2611.01560}, 2026.

\bibitem[Zhang et~al.(2026{\natexlab{a}})Zhang, Ge, Yoo, Fu, Yang, Liu,
  Saravanan, Yin, Yu, Niu, Wang, Herzig, Goldberg, Bai, Chan, Stoica, Kanazawa,
  Lei, Feng, and Darrell]{zhang2026rats}
Junyi Zhang, Jiaxin Ge, Hanjun Yoo, Letian Fu, Zihan Yang, Yaowei Liu, Raj
  Saravanan, Shaofeng Yin, Justin Yu, Dantong Niu, Zirui Wang, Roei Herzig, Ken
  Goldberg, Yutong Bai, David~M. Chan, Ion Stoica, Angjoo Kanazawa, Jiahui Lei,
  Haiwen Feng, and Trevor Darrell.
\newblock Playful agentic robot learning.
\newblock \emph{arXiv preprint arXiv:2606.19419}, 2026{\natexlab{a}}.

\bibitem[Zhang et~al.(2026{\natexlab{b}})Zhang, Niu, Liu, Guo, Xu, Hu, and
  Ding]{zhang2026diffuview}
Kaizhao Zhang, Tian Niu, Tianyu Liu, Chenen Guo, Zijun Xu, Qingda Hu, and
  Wenchao Ding.
\newblock {DiffuView}: Multi-view diffusion pretraining for 3d-aware robotic
  manipulation.
\newblock In \emph{IEEE/CVF Conference on Computer Vision and Pattern
  Recognition (CVPR)}, 2026{\natexlab{b}}.

\bibitem[Zhang et~al.(2026{\natexlab{c}})Zhang, Zhang, Gao, Li, Liu, Zhu, Qiu,
  Yan, Liu, Tang, Fang, Nie, Wei, Wang, Ding, and Yu]{zhang2026harnessvla}
Yixian Zhang, Huanming Zhang, Feng Gao, Xiao Li, Zhihao Liu, Chunyang Zhu,
  Jiaxing Qiu, Yuchen Yan, Jiyuan Liu, Wenhao Tang, Zhengru Fang, Yi Nie,
  Changxu Wei, Yu Wang, Wenbo Ding, and Chao Yu.
\newblock Harness vla: Steering frozen vlas into reliable manipulation
  primitives via memory-guided agents.
\newblock \emph{arXiv preprint arXiv:2607.08448}, 2026{\natexlab{c}}.

\bibitem[Zhao et~al.(2026)Zhao, Guo, Mei, Zhu, Zhang, Cao, Yu, He, Jiang, and
  Liu]{zhao2026atlasvla}
Guiyu Zhao, Longteng Guo, Yanghong Mei, Zilin Zhu, Yu Zhang, Bin Cao, Mingming
  Yu, Xingjian He, Jie Jiang, and Jing Liu.
\newblock {AtlasVLA}: Persistent world-ego state modeling for
  vision-language-action models.
\newblock \emph{arXiv preprint arXiv:2608.06729}, 2026.

\bibitem[Zhou et~al.(2025)Zhou, Xu, Tie, Chen, Zhang, Chu, Zhou, and
  Sun]{zhou2025liberopro}
Xueyang Zhou, Yangming Xu, Guiyao Tie, Yongchao Chen, Guowen Zhang, Duanfeng
  Chu, Pan Zhou, and Lichao Sun.
\newblock {LIBERO-PRO}: Towards robust and fair evaluation of
  vision-language-action models beyond memorization.
\newblock \emph{arXiv preprint arXiv:2510.03827}, 2025.

\bibitem[Zhou et~al.(2026)Zhou, Xu, Li, Tie, Wu, Huang, Zhang, Yuan, Wu, Chu,
  Zhou, Sun, and Chen]{zhou2026liberopro2}
Xueyang Zhou, Yangming Xu, Qianjiang Li, Guiyao Tie, Junhao Wu, Ziheng Huang,
  Guowen Zhang, Zhijie Yuan, Di Wu, Duanfeng Chu, Pan Zhou, Lichao Sun, and
  Yongchao Chen.
\newblock {LIBERO-Pro2.0}: A benchmark for evaluating the robustness of
  embodied foundation models.
\newblock \emph{arXiv preprint arXiv:2610.24587}, 2026.

\bibitem[Zhu et~al.(2020)Zhu, Wong, Mandlekar, Mart{\'i}n-Mart{\'i}n, Joshi,
  Lin, Maddukuri, Nasiriany, and Zhu]{zhu2020robosuite}
Yuke Zhu, Josiah Wong, Ajay Mandlekar, Roberto Mart{\'i}n-Mart{\'i}n, Abhishek
  Joshi, Kevin Lin, Abhiram Maddukuri, Soroush Nasiriany, and Yifeng Zhu.
\newblock robosuite: A modular simulation framework and benchmark for robot
  learning.
\newblock \emph{arXiv preprint arXiv:2009.12293}, 2020.

\end{thebibliography}
}
\end{document}